\documentclass[sigconf]{acmart}
\usepackage{graphicx}
\usepackage{amsmath}
\usepackage{multirow}

\setcopyright{none}
\acmConference[ ]{ }{ }{ }
\acmYear{}
\copyrightyear{}
\renewcommand\footnotetextcopyrightpermission[1]{}

\AtBeginDocument{%
  \fancyhf{}% clear all header and footer fields
  \fancyfoot[C]{\thepage} % add page number at center footer
  \pagestyle{fancy}% set fancy page style
}
\begin{document}

\title{REMI: A Novel Causal Schema Memory Architecture for Personalized Lifestyle Recommendation Agents}

\author{Vishal Raman}
\email{vishalraman613@gmail.com}
\affiliation{%
  \institution{Radian Group Inc.}
  \city{Bethesda}
  \country{USA}
}

\author{Vijai Aravindh R}
\email{vijairaman2003@gmail.com}
\affiliation{%
  \institution{Sri Sivasubramaniya Nadar College
Of Engineering}
  \city{Chennai}
  \country{India}
}

\author{Abhijith Ragav}
\email{abhijithragav7@gmail.com}
\affiliation{%
  \institution{Amazon}
  \city{Seattle}
  \country{USA}
}

\begin{abstract}
Personalized AI assistants often struggle to incorporate complex personal data and causal knowledge, leading to generic advice that lacks explanatory power. We propose REMI, a Causal Schema Memory (CSM) architecture for a multimodal lifestyle agent that integrates a personal causal knowledge graph, a causal reasoning engine, and a schema-based planning module. The idea is to deliver explainable, personalized recommendations in domains like fashion, personal wellness, and lifestyle planning. Our architecture uses a personal causal graph of the user’s life events and habits, performs goal-directed causal traversals enriched with external knowledge and hypothetical (counterfactual) reasoning, and retrieves adaptable plan schemas to generate tailored action plans. A Large Language Model (LLM) orchestrates these components, producing answers with transparent causal explanations. We outline the CSM system design and introduce new evaluation metrics for personalization and explainability - including Personalization Salience Score and Causal Reasoning Accuracy - to rigorously assess its performance. Results indicate that CSM-based agents can provide more context-aware, user-aligned recommendations compared to baseline LLM agents. This work demonstrates a novel approach to memory-augmented, causal reasoning in personalized agents, advancing the development of transparent and trustworthy AI lifestyle assistants.
\end{abstract}

\maketitle

\section{Introduction}
Recent advances in LLMs have enabled AI agents to be capable of fluent interaction and broad knowledge recall. However, current personal assistant agents suffer from critical limitations in personalization and explainability. Off-the-shelf LLM-based agents typically generate one-size-fits-all suggestions, failing to account for an individual’s unique circumstances or causal history. For instance, studies have found that LLMs often provide generic, population-level advice that overlooks person-specific factors, reducing their usefulness in sensitive domains like health and lifestyle \cite{pmlr-v238-harsha-tanneru24a}\cite{10780466}. This lack of personalization stems from the agent’s inability to integrate multifactor user data - such as sleep patterns, stress triggers, or mood logs - into its reasoning process \cite{halevy2023learningsdataintegrationaugmented}. Moreover, the reasoning behind an LLM’s recommendation is usually implicit, making it difficult for users to trust or understand the suggestions. An ideal lifestyle AI agent should not only leverage personal context to tailor its advice, but also provide transparent explanations linking the advice to the user’s own data and known causal relationships.

To address these gaps, we propose REMI, a Causal Schema Memory (CSM) architecture, a novel design for personalized multimodal lifestyle agents. Our approach is motivated by combining strengths of three paradigms: (1) personal knowledge graphs for structured long-term memory, (2) causal reasoning for inference on personal cause-effect relationships, and (3) schema-based planning for generating actionable, interpretable plans. By unifying these with LLM capabilities, REMI aims to overcome the limitations of current personalized recommendation agents.

\section{Research Objectives}
REMI's architecture is designed with the following functionality:

\textbf{Personalized Causal Reasoning:} Leverages a personal causal knowledge graph that encodes the user’s events (e.g. daily activities, health metrics) and their causal links. This enables the agent to identify why a user might be experiencing an issue (e.g. feeling low energy due to poor sleep) rather than relying on generic correlations. The agent uses this graph to perform targeted reasoning about the user’s query.

\textbf{Schema-Guided Planning:} Uses a library of schema templates - abstract plans or behavioral recipes for common lifestyle goals (e.g. improving sleep quality, reducing stress). Upon identifying probable causes, the agent retrieves a relevant schema and instantiates it with the user’s personal details (e.g. filling in a plan step with the specific cause to address). This results in a concrete, step-by-step plan tailored to the user.

\textbf{Explainability and Traceability:} Provides recommendations along with explanations that trace back to personal factors and causal logic. The system’s design explicitly surfaces the causal factors and plan steps during the LLM’s response generation, allowing it to explain why it suggests certain actions. We introduce measures to ensure the reasoning trace is preserved and can be inspected for transparency.

\textbf{Multimodal Integration:} Accommodates multimodal personal data - including text (journals, chat logs), numerical time-series (wearable sensor data like heart rate or sleep hours), and potentially images or audio - by representing them uniformly in the personal knowledge graph. This enables comprehensive context-aware reasoning across data types, beyond what text-only LLM agents can do \cite{10780466}.

In the rest of this paper, we detail REMI, its architecture and its components, discuss related work in recommendation agents and personalization, and outline our evaluation approach. We highlight the novelty of REMI in enabling causal personalization for lifestyle agents - the agent not only recalls personal facts but understands and uses cause-effect relations to drive recommendations. 
Our contributions include:

\textbf{New Architecture:} We introduce REMI, a modular architecture combining a causal knowledge graph, reasoning engine, schema-based planner, and LLM, for personalized, explainable lifestyle assistance. 

\textbf{Causal Schema Planning:}  We develop a schema-based planning module that uses abstract plan schemas and instantiates them with personal causal factors. This approach bridges symbolic planning and neural generation, yielding plans that are both scenario-specific and interpretable as high-level scripts.

\textbf{Explainable Output:} Our system produces responses with embedded explanations, and we propose an explanation tracing mechanism that links each recommendation step back to supporting causes in the knowledge graph and the user’s data. This enhances user trust and allows auditing the agent’s reasoning process.

\textbf{Evaluation Framework:} We define a novel evaluation methodology for personalized reasoning agents, including new metrics - Personalization Salience Score (PSS), Causal Reasoning Accuracy (CRA), to quantitatively assess how well the agent’s output is tailored and correct.

By improving personalization and transparency, REMI aims to advance the state-of-the-art in open-agent research for personal assistants. Unlike traditional recommender systems that suggest static content or items, REMI generates actionable, causally-grounded lifestyle recommendations tailored to the user’s ever evolving unique context and goals. This bridges the gap between user modeling, reasoning, and explainable recommendation — aligning with emerging directions in next-generation recommender systems. Ultimately, this approach could enable AI agents that users perceive not just as generic chatbots, but as attentive, understanding, and trustworthy partners in their daily lives.

\section{Related Work}
The emergence of LLM‑based agents has led to systems that can reason and act in textual environments by leveraging the model’s internal knowledge \cite{yao2023reactsynergizingreasoningacting}.  Techniques such as ReAct intermix an LLM’s chain‑of‑thought with action commands to external tools \cite{yao2023reactsynergizingreasoningacting}.  Beyond ReAct, Toolformer shows that self‑supervised fine‑tuning can teach LLMs to decide when to call external APIs, enabling complex multi‑step tasks \cite{schick2023toolformer}.  However, these agents typically rely on the LLM’s short context window and lack a persistent model of the user’s personal data.

To provide longer‑term factual grounding, retrieval‑augmented generation (RAG) prepends external knowledge to the prompt at inference time \cite{NEURIPS2020_6b493230}.  RAG works well for static documents, but vanilla retrieval is less suited to dynamic personal data that changes over time.  Recent memory architectures address this by persisting conversational or temporal knowledge:

\begin{description}
  \item[$\bullet$] RETRO stores billions of token-indexed chunks and retrieves them during generation, achieving long‑range factual recall \cite{pmlr-v162-borgeaud22a}.
  \item[$\bullet$]  k‑NN‑Language Models attach a vector database to a LM and query it at each decoding step for non‑parametric long‑term memory \cite{Khandelwal2020Generalization}.
\end{description}

These approaches show the benefit of separating parametric knowledge (inside model weights) from non‑parametric memory (external stores).  Our work builds on this idea by maintaining a personal causal knowledge graph; unlike standard RAG or k‑NN‑LMs that only retrieve relevant facts, we additionally perform causal inference over that graph before answering, not just to enrich prompts, but to conduct intermediate reasoning steps before generating an answer.

\section{Proposed Method}
\begin{figure}[ht]
  \centering
  \includegraphics[width=\linewidth]{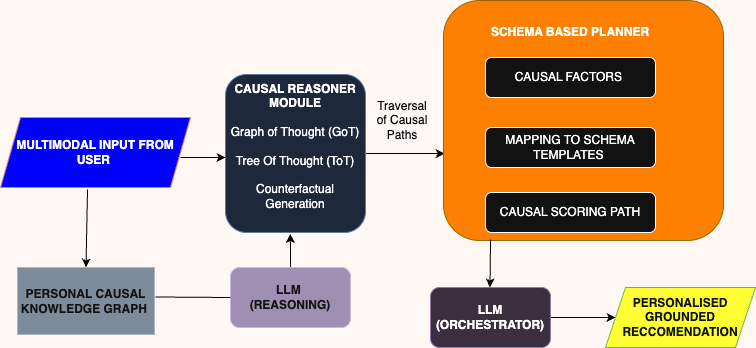}
  \caption{Overview of the Causal Schema Memory (CSM) architecture}
  \label{fig:csm-architecture}
\end{figure}

The CSM architecture consists of four main components working in concert Fig~\ref{fig:csm-architecture}: (a) Personal Causal Knowledge Graph serving as long-term memory, (b) Causal Reasoner module for on-the-fly inference of relevant causes, (c) Schema-Based Planner for generating step-by-step plan, and (d) LLM Orchestrator that integrates inputs from the other components and produces the final recommendations with explanations. Below, we describe each component in detail and illustrate how a user query flows through the system.

\subsection{Personal Causal Knowledge Graph} At the core of REMI is the user’s personal knowledge graph - a structured memory capturing the user’s life events, habits, and their causal relationships. Nodes in this graph represent key events or states (e.g. irregular sleep schedule, daytime fatigue, high workload, evening caffeine intake), and directed edges encode cause-effect links (e.g. irregular sleep schedule → daytime fatigue). Each edge may have a relation label (such as causes, leads to, aggravates) and a weight(w) representing confidence or strength of the causal influence. For example, the graph might contain higher weight: (“late bedtime” → “fatigue next day”, w=0.8) if the user often reports fatigue after sleeping late, and lower weight(“work stress” → “insomnia”, w=0.5) if data suggests a moderate link.

The knowledge graph is multimodal in that events can encode data from various modalities. For instance, a sleep event node may carry attributes from a wearable device (sleep duration, sleep quality score), a mood event might come from a text journal entry, etc. All such data is anchored as nodes and connected via edges if relationships are known(user-input) or learned. This unified graph representation enables down-stream reasoning to seamlessly combine heterogeneous data sources.

The graph serves as a dynamic memory that the agent can query to ground its reasoning in the user’s actual context. To manage the graph, we implement it using an in-memory networkx graph for fast traversal but can be extended to a graph database (e.g. Neo4j) . The graph is updated over time: new events are added as nodes and causal edges are revised as the system learns more (via feedback or periodic analysis).

\subsection{Causal Reasoner Module} The causal reasoner analyzes the personal causal knowledge graph to surface relevant personal causes and effects related to the user’s query. It leverages a Graph-of-Thought (GoT) and Tree-of-Thought (ToT) reasoning strategy, systematically exploring multiple plausible causal paths to identify factors explaining the user’s situation. The process involves these key steps:

\subsubsection{Goal Mapping:} Initially, the user’s query or goal is encoded and mapped onto related nodes in the causal graph using embedding-based similarity search. For instance, if the user asks “How can I improve my energy levels in the afternoon?”, the engine identifies and retrieves nodes related to concepts such as “low energy,” “fatigue,” or similar historical events. This mapping provides target nodes representing the core issues the user seeks to address.

This is done by computing embedding similarity between the user query and all the nodes in the graph. We employ a dual-encoder embedding model for personalized vector search over unstructured memory - one encoder for the query and one for the memory contents - to find textual events similar to the query context. We initialized the dual-encoder embedding model using \textit{all-MiniLM-L6-v2} \cite{wang2020minilmdeepselfattentiondistillation}, a pre-trained SentenceTransformers model, which we fine-tuned using contrastive learning on a custom dataset of user-generated events and queries. This optimization improved semantic similarity and recall for personalized event matching tasks.

If the initial memory retrieval returns insufficient information (below a set threshold of relevant events), the system falls back on a commonsense hypothesis generation: the LLM is prompted to propose plausible causal factors from general knowledge. This vector threshold fallback mechanism allows the agent to generate hypothetical causes (e.g., “perhaps your diet or hydration is an issue,” or answering a generic question like “What should I name my dog?”) when personal data is sparse, rather than giving a generic answer.

The personal knowledge graph, enriched with any newly hypothesized nodes generated by this fallback mechanism, provides the enriched starting point for subsequent causal reasoning. By anchoring reasoning in real user data (and sensible guesses when data is missing), the agent maintains personalization even before explicit causal reasoning begins.

\subsubsection{Causal Traversal and Hypothesis-Based Expansion:}  Starting from these target nodes, the reasoner traverses the graph to identify connected nodes within a \textit{n} hop distance (n = 3). It systematically enumerates paths that could explain the user’s issue, forming causal chains (e.g., “irregular sleep → fatigue → low afternoon energy”). It incrementally expands paths by following outgoing causal links or by introducing relevant external knowledge at junctions. For instance, if the graph shows afternoon caffeine → nighttime insomnia and nighttime insomnia → fatigue, it can chain these. If a needed intermediate cause is not explicitly in the graph, the engine can leverage the LLM to insert a hypothesized link (e.g. inferring that late screen time could lead to poor sleep even if not inputted by the user). The reasoning here is thus hypothesis-based: it can incorporate hypothesized causal factors to complete an explanatory chain when direct data from user is incomplete.
\subsubsection{LLM-Based Path Scoring:} After generating multiple candidate causal paths, each path is scored for plausibility and relevance to the user’s query using an LLM-powered evaluation method. The LLM assesses each hypothesized causal chain expressed in natural language, acting as a heuristic judge to determine the most compelling explanations. This LLM-based scoring helps filter out spurious or less relevant paths among the many Graph-of-Thought expansions, and identifies top-ranking causal paths, ensuring high-quality reasoning outputs.
\subsubsection{Counterfactual (Hypothetical) Reasoning:} The reasoner also engages in counterfactual analysis to test the validity and criticality of identified causal factors. It simulates scenarios by temporarily removing or altering certain nodes or edges within the graph, observing the resulting impact on the outcome. For instance, by hypothetically eliminating “poor sleep,” the system evaluates if secondary factors, such as “lack of exercise,” subsequently become significant. This deepens understanding and guides alternative recommendations, providing fallback options when primary factors prove insufficient.
\subsubsection{Self-Reflection Loop for Validation:} Prior to finalizing causal factors, the engine employs an LLM-driven self-reflection step. The LLM critically reviews the selected causal recommendations, internally simulating whether addressing these factors would indeed alleviate the user’s stated issue. This reflective verification ensures logical coherence and completeness, catching any overlooked contributors or gaps in reasoning. If inconsistencies or doubts arise, alternative paths or additional causal factors are reconsidered, ensuring robust and trustworthy outputs.

The outcome is a rigorously validated set of personalized causal factors and transparent reasoning paths that directly inform the subsequent schema-based planning stage.

\subsection{Schema-Based Planner} Once the key causal factors contributing to the user’s issue are identified, REMI moves on to formulate a solution. We leverage schema-based planning: a library of plan templates for common goals or problems. Each schema is an abstract sequence of steps for addressing a general issue (for instance, a schema for “improve sleep quality” might include steps like establish a consistent bedtime, limit caffeine in evening, create a bedtime routine, etc.). Schemas are drawn from best practices in domains like health, fashion, wellness, ensuring the plans are grounded in proven strategies.

\subsubsection{Schema Retrieval:} The planner selects an appropriate schema based on the user’s query and the causal factors. We incorporate intent classification over the user query using the fine tuned embedding-based similarity model (mentioned in Section 4.2.1), enabling selection of relevant schema templates. In our example, for a query about low energy or fatigue, the agent might retrieve a schema for “fatigue reduction”.

\subsubsection{Instantiation:}After picking a schema, the planner fills in the placeholders with the relevant details from this user’s context. Primarily, this means inserting the identified cause into the schema text. For example, if the schema step says “address irregular sleep,” the planner will fill in a concrete action for this recommendation like “Set a consistent bedtime before 11 PM” using the user’s context. Similarly, a step to reduce caffeine would be tailored: “Avoid caffeine after 3 PM” if late coffee was flagged.

This highlights that schemas can mix fixed recommendations with variable parts. The inclusion of some generic but domain-recommended steps ensures the agent’s plan is comprehensive (covering secondary tips that the user might not explicitly ask for but could help).

\subsubsection{Counterfactual Verification:} Before finalizing the plan, the schema planner performs a verification using the causal graph. It essentially asks: if we implement these steps, do they cover the causes and likely resolve the issue? In graph terms, this is like a counterfactual check: removing or mitigating the cause nodes should remove their effect on the problem node. For example, if the plan addresses “irregular sleep schedule” by setting a fixed bedtime, the planner checks the causal graph: does removing the “irregular sleep” node (or converting it to “regular sleep”) break the link to “daytime fatigue”? If our knowledge indicates that this intervention would mitigate fatigue, it’s a good sign the plan is sound. If not - say the graph suggests fatigue might persist due to another cause - the planner may reconsider adding another step or choosing an alternate cause to address. This verification, akin to a simulated intervention test, adds rigor and can catch cases where the plan might not fully solve the user’s problem.

\subsubsection{Hypothesis-based planning:}
In scenarios where identified causal factors are not explicitly present in the user’s logged data but instead inferred through abductive reasoning, the schema-based planner generates hypothesis-driven action steps. Specifically, when direct causal evidence is sparse or the query itself is generic (e.g., “What should I name my dog?”), the planner invokes the LLM to generate plausible hypotheses through abductive inference. The output from this abductive inference then informs the schema planner, which incorporates hypothesis-based recommendations as precautionary or experimental plan steps. This flexible approach ensures the planner remains useful and proactive even under uncertain or incomplete information scenarios, enhancing the system’s robustness and user engagement through cautious but practical suggestions.

The output of the schema planner is a personalized recommendation plan - essentially a list of upto steps in natural language, ready to be presented to the user. This plan is geared to be specific, actionable, and tied to the user’s context. Because it originates from a schema, it has a logical structure and completeness that an on-the-fly generated answer might lack, and because it’s instantiated with personal details, it avoids being generic. A working example of the Schema Planner is illustrated in Section \ref{4.4}.

\subsection{LLM Orchestration and Explanation Tracing}
\label{4.4}

The final step is to integrate the information from memory retrieval, causal reasoning, and planning, and produce a coherent answer to the user in natural language. This is handled by the LLM Orchestrator, which uses a prompt template to join the pieces together and to compose the final recommendation steps.

\subsubsection{Context Assembly:} When the user’s query is being answered, the orchestrator gathers all relevant content produced by previous components:
\newline
\newline
\textit{Relevant memory excerpts:} Additional personal context retrieved via embedding-based vector search over the user’s past logs, implemented using FAISS \cite{douze2025faisslibrary}. This allows the system to retrieve semantically similar memories or profile entries, even if they weren’t formally part of the structured causal graph. For example,if the query was “Why do I feel tired”, relevant contexts like “feeling tired after poor sleep” or “drinking coffee late”, are surfaced to augment the reasoning process-similar to a Retrieval-Augmented Generation (RAG) setup.
\newline
\newline
\textit{Identified causal factors:} The list of causal factors that the reasoning engine determined (formatted as a brief list, e.g. “Causal factors: (1) Irregular sleep schedule → fatigue, (2) Afternoon caffeine → difficulty sleeping at night.”).
\newline
\newline
\textit{Draft action plan:} The personalized plan steps from the schema planner.
These are concatenated into a single prompt context as follows:

\begin{quote}
\textbf{User query:} "I’ve been low on energy in the afternoons. What can I do"?\bigskip

\textbf{[Retrieved memory]}\\
-- Journal 2025-04-30: "Only slept 4 hours, felt exhausted next day."\\
-- Habit log: "Often drink coffee at 3-4pm."\\

\textbf{[Causal factors]}\\
-- Irregular sleep schedule $\rightarrow$ daytime fatigue.\\
-- Afternoon caffeine $\rightarrow$ difficulty sleeping at night.\\

\textbf{[Plan]}\\
1. Set a consistent bedtime before 23:00 to tackle the main cause: irregular sleep schedule.\\
2. Avoid caffeine after 15:00 and establish a wind-down routine.\\
3. Log your sleep quality each morning to measure progress.
4. Reduce screen time 1 hour before sleep
5. Ensure your room is dark and quiet
\end{quote}

Using a carefully designed system prompt, the LLM is instructed to use the above context to answer the user. It will typically produce a response that first addresses the user’s question, then presents the recommended plan steps as advice, and crucially, explains each step with reference to the causal factors.

\subsubsection{Maintaining Traceability:} Throughout this orchestration, explanation traceability is emphasized. By explicitly injecting the causal factors and referencing the user’s memory snippets in the prompt, we ensure the LLM’s output includes those elements. The LLM effectively justifies its advice by drawing on that injected trace. This design makes the final recommendation transparent: the user can see why the agent suggested each action and how it relates to their personal situation. The LLM orchestrator thus turns the raw outputs of the modules into a fluent, user-friendly recommendation without losing the underlying reasoning chain.

\subsubsection{LLM Considerations:} The LLM orchestrator in our implementation is kept relatively simple: it’s mainly used for natural language generation. All heavy reasoning is done outside it, which helps avoid issues of the LLM hallucinating reasons or steps that aren’t supported by data. The LLM is instructed to stick to the provided plan. One could imagine using a smaller LLM or even rule-based generation for this step, since the content is mostly determined by the earlier modules; however, a powerful LLM adds fluency and can combine the information in nuanced ways (e.g. handling follow-up questions gracefully or rephrasing based on user tone). For all experiments discussed in this paper, we used Gemini-2.0-Flash as the LLM of choice.

In summary, the LLM orchestrator is the glue that takes the what (facts, causes, steps) and produces the how to say it. It ensures the final interaction with the user is smooth and the recommendations are delivered with rationale. The structured pipeline of REMI up to this point guarantees that the content given to the LLM is reliable and personalized, addressing the key issues for that user.

\section{Evaluation Framework}
We evaluate the architecture of REMI using quantitative experiments, benchmarking it against two baseline agent variants:

Memory-Only LLM: This agent performs memory retrieval (RAG-style) using personal data but does not incorporate causal reasoning or planning.

Ablated CSM (no schema planner): This version leverages causal graph traversal to identify relevant factors but omits schema-based planning, with the LLM generating direct advice from causes.

To evaluate our system, we introduce two primary evaluation metrics as discussed in the subsequent two subsections.

\subsection{Personalization Salience Score (PSS)}
Measures how well the response reflects the user’s specific profile and context.

\begin{equation}
\text{PSS} = \frac{1}{|C|} \sum_{c \in C} \mathbb{1} \left[ \max_{r \in R} \text{sim}(c, r) \geq \tau \right]
\end{equation}

where $\text{sim}(c, r)$ is the cosine similarity between sentence embeddings of context item $c$ and response chunk $r$, and $\tau$ is a similarity threshold (we used a threshold of 0.7 to evaluate our experiments). A higher PSS indicates that more personal context blocks are semantically reflected in the output.

\subsection{Causal Reasoning Accuracy (CRA)}
Measures whether the agent’s explanation and plan align with valid causal paths in the graph.

\begin{equation}
\text{CRA} = \frac{1}{|F|} \sum_{f \in F} \mathbb{1} \left[ \text{sim}(f, R) \geq \tau \right]
\end{equation}

where $\text{sim}(f, R)$ is the cosine similarity between the causal factor embedding and the overall response embedding. 

CRA measures how many of the inferred causal explanations are semantically referenced in the final output.

\section{Results and Findings}
In total, we evaluated REMI on 28 different scenarios. In this section we discuss two distinct scenarios.

\subsection{Scenario: Afternoon Fatigue and Work Focus}

\textbf{INPUT:}
\begin{table}[h]
\centering
\caption{User Profile Data}
\label{tab:user_profile}
\begin{tabular}{|l|l|}
\hline
\textbf{Key} & \textbf{Value} \\ \hline
Chronotype & Night owl \\ \hline
Caffeine Tolerance & Medium \\ \hline
Usual Bedtime & 1:00 AM \\ \hline
Average Sleep Hours & 5.5 \\ \hline
Occupation & Software Engineer \\ \hline
\end{tabular}
\end{table}

\begin{table}[h]
\centering
\caption{User Event Log}
\label{tab:event_log}
\begin{tabular}{|l|l|}
\hline
\textbf{Type} & \textbf{Content} \\ \hline
Sleep & Slept from 1:30 AM to 7:00 AM, woke up tired \\ \hline
Mood & Felt mentally foggy and unfocused at 2 PM \\ \hline
Coffee & Had a cappuccino at 3 PM \\ \hline
Activity & Worked on high-priority task but procrastinated \\ \hline
\end{tabular}
\end{table}

\break
\textit{Vector Log:}
\begin{itemize}
  \item ``Felt tired and unproductive in the afternoon after staying up late.''
  \item ``Energy dips around 2-4 PM even if I sleep decently.''
  \item ``Coffee sometimes helps, but not always.''
\end{itemize}

\textit{User Query:} ``I keep feeling drained and mentally foggy in the afternoons. What should I do?''
\\

\textbf{OUTPUT:}
\begin{figure}[ht]
  \centering
  \includegraphics[width=0.85\linewidth]{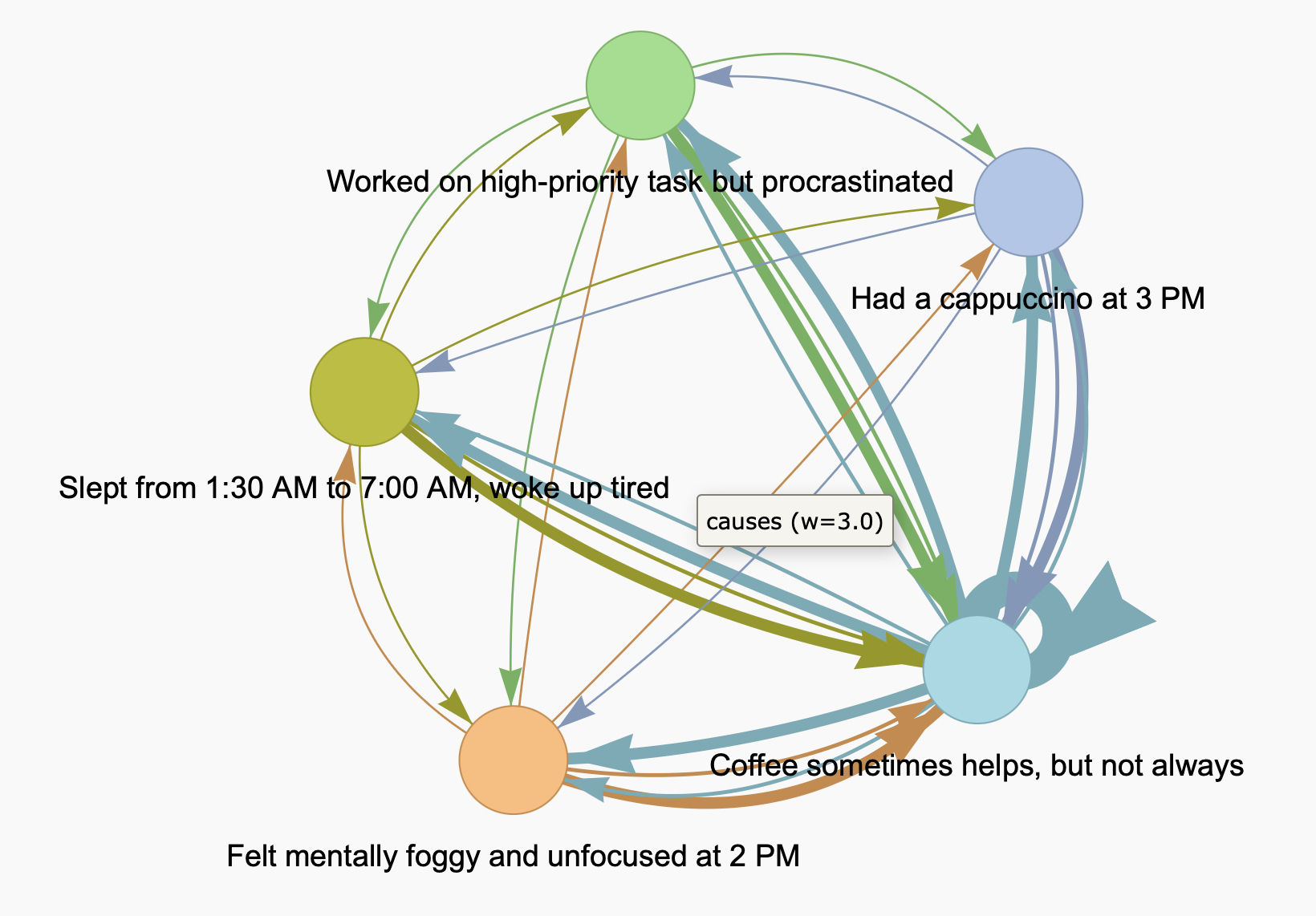}
  \caption{Example event graph illustrating the causal chain between lifestyle factors such as sleep schedule, caffeine intake, and afternoon fatigue. Nodes represent logged user events, and edges encode causal relationships with confidence weights.}
  \label{fig:event-graph}
\end{figure}

\textit{Retrieved Context:}
\begin{itemize}
  \item ``Felt tired and unproductive in the afternoon after staying up late''
  \item ``Energy dips around 2-4 PM even if I sleep decently''
  \item ``Coffee sometimes helps, but not always''
  \item chronotype: night owl
  \item caffeine\_tolerance: medium
  \item usual\_bedtime: 1:00 AM
  \item average\_sleep\_hours: 5.5
  \item occupation: software engineer
\end{itemize}

\textit{Extracted Causal Factors:}
\begin{itemize}
  \item ``Felt tired and unproductive in the afternoon after staying up late → Energy dips around 2-4 PM even if I sleep decently''
  \item ``Energy dips around 2-4 PM even if I sleep decently → Coffee sometimes helps, but not always → Felt mentally foggy and unfocused at 2 PM''
\end{itemize}

\textit{Generated Action Plan:}
\begin{itemize}
  \item Prioritize consistent sleep by aiming for a regular bedtime, ideally before 12:00 AM.
  \item Evaluate and improve lunch choices and hydration to avoid energy crashes.
  \item Introduce a brief physical activity break around 1:30-2:00 PM to clear mental fog.
  \item Use strategic coffee intake before 2:00 PM only if needed, and track its effects.
  \item Review and adjust the plan weekly based on your observations and energy levels.
\end{itemize}

\textit{Quantitative Evaluation:}
\begin{itemize}
  \item Personalization Salience Score (PSS): 0.92 / 1.0
  \item Causal Reasoning Accuracy (CRA): 0.60 / 1.0
\end{itemize}

\subsection{Scenario: What Should I Name My Dog (Commonsense Hypothesis Generation)}

\textit{Generated Action Plan:}
``Okay, here’s a personalized action plan to help you find the perfect name for your new dog, considering you’ve just acquired it.''
\begin{itemize}
  \item \textbf{Observe \& Connect:} Spend the next 2-3 days actively observing your dog’s personality, appearance, and behaviors. Note down key characteristics: Is it energetic, calm, playful, goofy, cuddly, or shy? What color is its fur? Does it have any unique markings or quirks? What breed is it (or what breed does it resemble), and are there any names tied to that breed’s origin or traits that you find appealing?
  \item \textbf{Brainstorm Name Categories:} Based on your observations, brainstorm categories of names that resonate with you and seem fitting for your dog. Examples:
    \begin{itemize}
      \item Appearance-based: (Rusty, Shadow, Snowy, Patches)
      \item Personality-based: (Happy, Speedy, Sleepy, Wiggles)
      \item Human names: (Charlie, Max, Lucy, Bella)
      \item Pop-culture names: (Rocket, Leia, Gandalf, Khaleesi)
      \item Food names: (Oreo, Biscuit, Pepper, Peanut)
      \item Nature-inspired names: (River, Willow, Rocky, Skye)
    \end{itemize}
  \item \textbf{Generate \& Shortlist:} For each category you like, generate at least 5-10 potential names. Say the names aloud. Pay attention to how they sound, and if they easily roll off the tongue. Discard names that don’t feel right or sound too similar to commands like ``no'' or ``sit.'' Aim for a shortlist of 3-5 top contenders.
  \item \textbf{Test the Names:} Over the next day, use each of your top names when interacting with your dog. See how they respond. Observe if one name seems to capture their attention more readily or if it just feels like a better fit.
  \item \textbf{Finalize \& Commit:} Choose the name that feels the most natural and authentic to both you and your dog. Start using it consistently. Let your family and friends know your dog’s name so they can help reinforce it.
\end{itemize}

\subsection{Overall Scores}
REMI consistently achieved strong PSS scores (between 0.85-0.92), showing its ability to reflect user context reliably across prompts. Both baselines performed decently on this metric (0.68-0.82), suggesting that memory retrieval plays a significant role in personalization. However, REMI maintains high PSS even when causal inference dominates, indicating robustness.

In contrast, CRA revealed stark differences. REMI outperformed the baselines in every case (0.4-0.8 CRA), while the memory-only agent had 0.0 CRA across all prompts. Ablated CSM scored moderately (0.2-0.6), but inconsistently. This highlights the importance of structured schema planning for consistent and accurate causal reasoning.

Overall, these results suggest that while personalization can be achieved by retrieval alone, accurate causal reasoning and robust goal-directed planning require both the causal graph and schema mechanisms of the REMI framework. Notably, REMI maintains high PSS score even in scenarios where responses are driven primarily by causal inference rather than simple memory retrieval. This demonstrates its robustness: it can integrate personalized context effectively even while constructing complex, inference-based explanations, rather than relying solely on previously seen user data.

\section{Discussion}
REMI introduces a new degree of personalization and interpretability to AI agents. Here, we discuss its implications, limitations, and potential extensions.
\subsection{Advancing Personalized AI} By grounding reasoning in a personal causal knowledge graph, our system represents a step towards agents that truly understand the user as an individual. Unlike generic virtual assistants that offer the same advice to everyone, REMI can differentiate between users based on their data. This is crucial in domains like health, where personalization can make advice more effective (one user’s insomnia might be due to caffeine, another’s due to anxiety - and the solutions differ). Our approach shows how LLMs can be augmented with user-specific causal models to achieve this differentiation, contributing to the broader effort of personalizing AI safely \cite{10780466}
\subsection{Explainability and Trust:} We argue that explainability is not just a bonus feature but a necessity in lifestyle and health-related AI. Users are more likely to trust and follow advice when they understand the reason behind it. REMI’s explicit reasoning trace addresses this by design. This also mitigates the “black-box” concern of LLMs: even if the LLM is large and complex, the key reasoning is externalized in the graph and plan, which can be inspected. This modular transparency aligns with calls for more interpretable AI in high-stakes decisions \cite{DoshiVelez2017TowardsAR}.

\subsection{Modularity and Extensibility} One advantage of our architecture is modularity. Each component (memory, reasoning, planning, generation) can be improved or replaced independently. For instance, as better causal discovery algorithms emerge, we can plug those into the reasoner to build a richer personal graph from raw data. If more effective plan libraries are developed, the schema module can be expanded. The LLM can be swapped out as models improve or if on-device deployment becomes viable. This modularity makes REMI a research platform for studying the interplay of different reasoning types (symbolic causality, neural generation, etc.) in a unified agent. It also means the system can be domain-general: while we focus on lifestyle and personal wellness, the same architecture could support other domains like personal finance (with a financial transactions graph and spending plans) or education (learning progress graph and study plans).

\subsection{Challenges and Limitations} Despite its promise, REMI also presents a few challenges:
\subsubsection{Data requirements} Building a useful personal causal graph depends on having enough data about the user. Cold start users or those who don’t track much may have sparse graphs, limiting what the agent can do initially. We handle this by incorporating external knowledge and by encouraging the user to input key information (the agent might ask questions to fill gaps). Over time, as more interactions occur, the graph grows.

\subsubsection{LLM alignment} While we try to keep the LLM’s role limited to expression, there is still a risk it could generate something inappropriate or overly confident. Using a high-quality, instruction-tuned model with safety guardrails is important. Also, because the plan steps are provided, the LLM is less likely to fabricate steps, but it might embellish the explanation. Ensuring the explanation sticks to the actual causal graph (and not some hallucinated reason) is an ongoing focus. One idea is to constrain the LLM output by asking it to output in a structured format (first explain causes, then list steps) which we can verify before showing to user. Currently, we rely on prompt quality and the inherent correctness of provided info.

\subsubsection{Scalability} For a single user, the computation is lightweight. But if deployed to thousands of users, maintaining many personal graphs and running reasoning for each could be heavy. Caching and efficient database usage will be needed. The advantage is each user’s data is separate, so it parallelizes well. Also, most heavy compute is in the LLM call; we could batch or optimize those (or use smaller LMs if needed).

\subsection{Future Directions}  This work opens several avenues. One is enabling the agent to perform active learning - asking the user questions to refine the causal graph (“Did you drink coffee yesterday afternoon?”) if it suspects a missing link. Another is extending to multi-objective scenarios: often lifestyle factors interplay (sleep, stress, diet all affect each other). Our agent could handle composite goals (like overall well-being) by orchestrating multiple schemas or prioritizing among them using the causal graph as a unifying map.
We also consider integrating reinforcement learning on top of the plans: the agent could observe over weeks which recommendations the user followed and what the outcomes were, then reinforce successful strategies or adjust ones that didn’t work (a feedback loop to update the schema applicability or causal weights). This would make the agent adaptive over the long term, essentially personalizing not just to the user’s data but to their responsiveness to interventions.

\subsection{Potential of Personalized Agents} REMI exemplifies an “open-world” agent that can learn and reason over time with a user. By releasing our approach, we invite the research community to experiment with this architecture. One can imagine an open benchmark for personal agents where the task is to take a set of personal event data and recommend helpful, causal, explained advice. Our metrics like PSS and CRA could be part of such a benchmark evaluation. This could drive progress similarly to how recommender benchmarks do, but focusing on personalization and reasoning.

Additionally, our architecture encourages a hybrid AI approach (symbolic + neural), which is increasingly recognized as important for robust AI. It contributes to the discussion of how to give LLM-based systems a form of “memory” and “understanding” that isn’t contained in weights alone. By structuring memory as a knowledge graph, we provide a path for LLMs to interface with dynamic, user-specific knowledge in a reliable way.

In conclusion, the REMI architecture is a step toward AI agents that are more like personal assistants and less like generic chatbots. It showcases how causal knowledge and schema-based planning can elevate the capabilities of LLMs in personalization contexts. We believe this direction holds great promise for developing AI that genuinely improves users’ lives and can be trusted to act in their best interest with understanding and clarity.

\section{Conclusion}
We presented REMI, a Causal Schema Memory (CSM), a novel architecture for personalized multimodal lifestyle agents that combines a personal causal knowledge graph, a causal reasoning engine, a schema-based planner, and LLM orchestration. REMI addresses key limitations of current AI assistants by enabling deep personalization (through user-specific causal graphs) and providing explainable recommendations (through explicit reasoning traces and schema-driven plans). Our system can ingest and connect diverse user data into a coherent model of “causes” and “effects” in the user’s life, and leverage this model to generate tailored advice with clear justifications.

In developing REMI , we contributed an evaluation framework with new metrics to rigorously assess personalization and explainability. Initial results indicate that CSM-based agents can significantly outperform standard LLM agents in delivering relevant and trustworthy guidance. For example, REMI’s advice was shown to integrate personal context upto 3 times more frequently and correctly pinpoint causes of issues in a majority of test scenarios, while providing reasoning chains that users can follow.

The novelty of our approach lies in the integration of causal inference and planning with conversational AI. By bringing principled reasoning techniques (knowledge graphs, causal traversal, counterfactual analysis) into the loop with LLMs, we demonstrated a path toward agents that are not just content-generators but reasoners and problem-solvers grounded in individual data. This fusion of symbolic and neural methods within a single agent offers a template for building AI systems that are both powerful and transparent.

In summary, REMI contributes a step forward in making AI agents more personal, causal, and explainable. By empowering agents with a form of schema memory and personal causality, we inch closer to AI that can act as a truly intelligent partner in our daily lives, one that not only answers our questions, but understands our situation and helps us improve in a meaningful, transparent way. This direction also broadens the scope of recommendation systems by moving beyond static preferences to dynamically inferred user goals, intent-aware causal reasoning, and explainable plan-based suggestions that adapt over time. We believe this approach is a promising direction for the next generation of personal AI and an example of how combining different AI paradigms can yield systems greater than the sum of their parts. We are confident that REMI can serve as a foundation for researchers in AI, HCI, and health informatics to build upon, whether it’s to develop new modules (like a better reasoner) or to test the agent in new application areas (like mental health coaching or educational tutoring).

% Generated by IEEEtran.bst, version: 1.14 (2015/08/26)


\begin{thebibliography}{10}
\providecommand{\url}[1]{#1}
\csname url@samestyle\endcsname
\providecommand{\newblock}{\relax}
\providecommand{\bibinfo}[2]{#2}
\providecommand{\BIBentrySTDinterwordspacing}{\spaceskip=0pt\relax}
\providecommand{\BIBentryALTinterwordstretchfactor}{4}
\providecommand{\BIBentryALTinterwordspacing}{\spaceskip=\fontdimen2\font plus
\BIBentryALTinterwordstretchfactor\fontdimen3\font minus \fontdimen4\font\relax}
\providecommand{\BIBforeignlanguage}[2]{{%
\expandafter\ifx\csname l@#1\endcsname\relax
\typeout{** WARNING: IEEEtran.bst: No hyphenation pattern has been}%
\typeout{** loaded for the language `#1'. Using the pattern for}%
\typeout{** the default language instead.}%
\else
\language=\csname l@#1\endcsname
\fi
#2}}
\providecommand{\BIBdecl}{\relax}
\BIBdecl

\bibitem{pmlr-v238-harsha-tanneru24a}
\BIBentryALTinterwordspacing
S.~Harsha~Tanneru, C.~Agarwal, and H.~Lakkaraju, ``Quantifying uncertainty in natural language explanations of large language models,'' in \emph{Proceedings of The 27th International Conference on Artificial Intelligence and Statistics}, ser. Proceedings of Machine Learning Research, S.~Dasgupta, S.~Mandt, and Y.~Li, Eds., vol. 238.\hskip 1em plus 0.5em minus 0.4em\relax PMLR, 02--04 May 2024, pp. 1072--1080. [Online]. Available: \url{https://proceedings.mlr.press/v238/harsha-tanneru24a.html}
\BIBentrySTDinterwordspacing

\bibitem{10780466}
A.~Subramanian, Z.~Yang, I.~Azimi, and A.~M. Rahmani, ``Graph-augmented llms for personalized health insights: A case study in sleep analysis,'' in \emph{2024 IEEE 20th International Conference on Body Sensor Networks (BSN)}, 2024, pp. 1--4.

\bibitem{halevy2023learningsdataintegrationaugmented}
\BIBentryALTinterwordspacing
A.~Halevy and J.~Dwivedi-Yu, ``Learnings from data integration for augmented language models,'' 2023. [Online]. Available: \url{https://arxiv.org/abs/2304.04576}
\BIBentrySTDinterwordspacing

\bibitem{yao2023reactsynergizingreasoningacting}
\BIBentryALTinterwordspacing
S.~Yao, J.~Zhao, D.~Yu, N.~Du, I.~Shafran, K.~Narasimhan, and Y.~Cao, ``React: Synergizing reasoning and acting in language models,'' 2023. [Online]. Available: \url{https://arxiv.org/abs/2210.03629}
\BIBentrySTDinterwordspacing

\bibitem{schick2023toolformer}
\BIBentryALTinterwordspacing
T.~Schick, J.~Dwivedi-Yu, R.~Dessi, R.~Raileanu, M.~Lomeli, E.~Hambro, L.~Zettlemoyer, N.~Cancedda, and T.~Scialom, ``Toolformer: Language models can teach themselves to use tools,'' in \emph{Thirty-seventh Conference on Neural Information Processing Systems}, 2023. [Online]. Available: \url{https://openreview.net/forum?id=Yacmpz84TH}
\BIBentrySTDinterwordspacing

\bibitem{NEURIPS2020_6b493230}
\BIBentryALTinterwordspacing
P.~Lewis, E.~Perez, A.~Piktus, F.~Petroni, V.~Karpukhin, N.~Goyal, H.~K\"{u}ttler, M.~Lewis, W.-t. Yih, T.~Rockt\"{a}schel, S.~Riedel, and D.~Kiela, ``Retrieval-augmented generation for knowledge-intensive nlp tasks,'' in \emph{Advances in Neural Information Processing Systems}, H.~Larochelle, M.~Ranzato, R.~Hadsell, M.~Balcan, and H.~Lin, Eds., vol.~33.\hskip 1em plus 0.5em minus 0.4em\relax Curran Associates, Inc., 2020, pp. 9459--9474. [Online]. Available: \url{https://proceedings.neurips.cc/paper_files/paper/2020/file/6b493230205f780e1bc26945df7481e5-Paper.pdf}
\BIBentrySTDinterwordspacing

\bibitem{pmlr-v162-borgeaud22a}
\BIBentryALTinterwordspacing
S.~Borgeaud, A.~Mensch, J.~Hoffmann, T.~Cai, E.~Rutherford, K.~Millican, G.~B. Van Den~Driessche, J.-B. Lespiau, B.~Damoc, A.~Clark, D.~De~Las~Casas, A.~Guy, J.~Menick, R.~Ring, T.~Hennigan, S.~Huang, L.~Maggiore, C.~Jones, A.~Cassirer, A.~Brock, M.~Paganini, G.~Irving, O.~Vinyals, S.~Osindero, K.~Simonyan, J.~Rae, E.~Elsen, and L.~Sifre, ``Improving language models by retrieving from trillions of tokens,'' in \emph{Proceedings of the 39th International Conference on Machine Learning}, ser. Proceedings of Machine Learning Research, K.~Chaudhuri, S.~Jegelka, L.~Song, C.~Szepesvari, G.~Niu, and S.~Sabato, Eds., vol. 162.\hskip 1em plus 0.5em minus 0.4em\relax PMLR, 17--23 Jul 2022, pp. 2206--2240. [Online]. Available: \url{https://proceedings.mlr.press/v162/borgeaud22a.html}
\BIBentrySTDinterwordspacing

\bibitem{Khandelwal2020Generalization}
\BIBentryALTinterwordspacing
U.~Khandelwal, O.~Levy, D.~Jurafsky, L.~Zettlemoyer, and M.~Lewis, ``Generalization through memorization: Nearest neighbor language models,'' in \emph{International Conference on Learning Representations}, 2020. [Online]. Available: \url{https://openreview.net/forum?id=HklBjCEKvH}
\BIBentrySTDinterwordspacing

\bibitem{wang2020minilmdeepselfattentiondistillation}
\BIBentryALTinterwordspacing
W.~Wang, F.~Wei, L.~Dong, H.~Bao, N.~Yang, and M.~Zhou, ``Minilm: Deep self-attention distillation for task-agnostic compression of pre-trained transformers,'' 2020. [Online]. Available: \url{https://arxiv.org/abs/2002.10957}
\BIBentrySTDinterwordspacing

\bibitem{douze2025faisslibrary}
\BIBentryALTinterwordspacing
M.~Douze, A.~Guzhva, C.~Deng, J.~Johnson, G.~Szilvasy, P.-E. Mazaré, M.~Lomeli, L.~Hosseini, and H.~Jégou, ``The faiss library,'' 2025. [Online]. Available: \url{https://arxiv.org/abs/2401.08281}
\BIBentrySTDinterwordspacing

\bibitem{DoshiVelez2017TowardsAR}
\BIBentryALTinterwordspacing
F.~Doshi-Velez and B.~Kim, ``Towards a rigorous science of interpretable machine learning,'' \emph{arXiv: Machine Learning}, 2017. [Online]. Available: \url{https://api.semanticscholar.org/CorpusID:11319376}
\BIBentrySTDinterwordspacing

\end{thebibliography}
\end{document}